\documentclass[11pt,a4paper]{article}
\usepackage[hyperref]{emnlp2020}
\usepackage{times}
\usepackage{latexsym}

\usepackage{microtype}
\usepackage{amsmath}
\usepackage{float}
\usepackage{lingmacros}
\usepackage{graphicx} % takes care of graphic including machinery
\usepackage{hyperref}
\usepackage{multirow}
\usepackage{threeparttable}
\usepackage{booktabs}
\usepackage{subcaption}
\usepackage{csquotes}
\usepackage{tcolorbox}

\newcommand*{\FirstIndent}{\hspace*{0.5cm}}%

\newcommand{\qgnumsample}[4]{
	\begin{displayquote}
		\textbf{Sentence #1:} #2\\
		\textbf{Question #1:} #3\\
		\textbf{Suggested answer #1:} #4
	\end{displayquote}
}

\aclfinalcopy % Uncomment this line for the final submission

\usepackage{fancyhdr}
\title{Automatically generating question-answer pairs for assessing basic reading comprehension in Swedish}

\author{Dmytro Kalpakchi \\
  Division of Speech, Music and Hearing \\
  KTH Royal Institute of Technology \\
  Stockholm, Sweden \\
  \texttt{dmytroka@kth.se} \\\And
  Johan Boye \\
  Division of Speech, Music and Hearing \\
  KTH Royal Institute of Technology \\
  Stockholm, Sweden \\
  \texttt{jboye@kth.se} \\}

\date{}

\begin{document}
\maketitle
\begin{abstract}
This paper presents an evaluation of the quality of automatically generated reading comprehension questions from Swedish text, using the Quinductor method. This method is a light-weight, data-driven but non-neural method for automatic question generation (QG). The evaluation shows that Quinductor is a viable QG method that can provide a strong baseline for neural-network-based QG methods.
\end{abstract}

% TODO: let Johan proofread the evaluation guidelines and put them in the article

\section{Introduction}
% \citet{snow2002reading} defines reading comprehension (RC) as ``the process of simultaneously extracting and constructing meaning through interaction and involvement with written language''. \citet{snow2002reading} elaborates that RC is in fact an interplay between the reader (their background knowledge, abilities, experiences, etc), the text to be comprehended and the activity (i.e., why the reading takes place, what consequences it entails, etc). From this complex and broad definition alone, it is apparent that in order to assess RC in general one needs to be able to generate a multitude of questions of various types (e.g.., multiple choice, cloze, summarize, etc) for texts in various genres.

In this article, we aim to establish a strong non-neural, but still data-driven, baseline for the automatic generation of reading comprehension questions (RC-QG) in Swedish. It is well-known that reading comprehension is a complex and multi-layered process, ranging from the simple decoding of individual words to advanced analytical tasks concerning the quality, veracity and purpose of whole texts \citep{alderson2000assessing,shaw2007examining}.
% \bibitem Alderson, C. (2000). Assessing Reading. Cambridge University Press.
% \bibitem Khalifa, H. & Weir, C. (2009). Examining Reading - Research and Practice in Assessing Second Language Reading. Cambridge University Press. 
In this work, we aim at only generating RC questions on the \emph{the information-locating level}, i.e.\ questions that facilitate the assessment of readers' ability to scan, locate and retrieve relevant information from a single text. This is the most basic level of RC, as identified by the PISA 2018 report \cite[p.34]{pisa2019}, but still a crucial reading ability in need of assessment. 

Our approach is data-driven, meaning that no handcrafted rules or linguistic expertise is required. Adapting to new kinds of questions is done via adding new text-question pairs to the training material. Furthermore, the generated questions and their respective answers (the {\em QA-pairs\/}) will quite faithfully reuse the wordings of the text, but can also use words and phrases that are not explicitly present in the text.
 %, although it does not paraphrase already available information.
Our system is completely open-source with source code available on GitHub (\url{https://github.com/dkalpakchi/swe_quinductor}).

\section{Related work}
To the best of our knowledge, the literature on reading comprehension question generation (RC-QG) particularly for Swedish is very limited. 
%In fact, the only two articles that we were able to find do not specify assessing reading comprehension as the primary goal of their QG system.
\citet{wilhelmsson2011automatic,wilhelmsson2012automatic} presented a system for generating questions using manually specified grammatical transformations (syntactic fronting of unbounded constituents and substitution of suitable question elements with question words). By design, their system is able to generate QA-pairs only using formulations that appear word-by-word in the text. The generated questions were limited to two categories, the first of which encompasses questions starting with ``vem'' (eng. ``who/whom''), ``vad'' (eng. ``what''), ``vilken'' (eng. ``which'') that would concern nominal constituents. The second category concerns questions to some adverbials.
%, i.e. NP adverbials predominantly referring time (replaced by ``n\"ar'' (eng. ``when'')), PP adverbials and some sub-clause types corresponding to adverbials. 
The author also presented a preliminary evaluation of the questions generated for ten random Wikipedia articles, but unfortunately neither specified the exact articles nor released the source code of his system, making a direct comparison impossible.

Lately, neural-network-based text generation methods have become very popular due their impressive results. Indeed, these methods have been applied to RC-QG as well, mostly for English (see e.g.\ \citet{liao2020probabilistically,dong2019unified}). Our goal here is not to compete with these approaches, but rather to present a more light-weight but still strong baseline to which neural methods can be compared, but which is also interesting and useful in its own right. 

%\citet{wilhelmsson2012automatic} essentially describes the same system as \citet{wilhelmsson2011automatic}, additionally surveying different kinds of questions that could be generated, the state of NLP techniques for doing that back in 2012, and potential applications for those kinds of questions.

\section{Data}
We have used the \emph{SweQUAD-MC} dataset \citep{kalpakchi-boye-2021-bert} consisting of texts and multiple-choice reading comprehension questions (MCQs) for the given texts. It was created by three paid linguistics students. They were instructed to formulate unambiguous questions, such that (1) the answer to the question appears verbatim in the text, (2) the question cannot be answered without reading the text (i.e., the text is {\em necessary\/}), and (3) the answer does not require extra knowledge not present in the text (i.e., the text is {\em sufficient\/}). These characteristics make the dataset suitable for assessing RC on the information-locating level.

\begin{figure}[!hbt]
	\centering
	\includegraphics[width=0.47\textwidth]{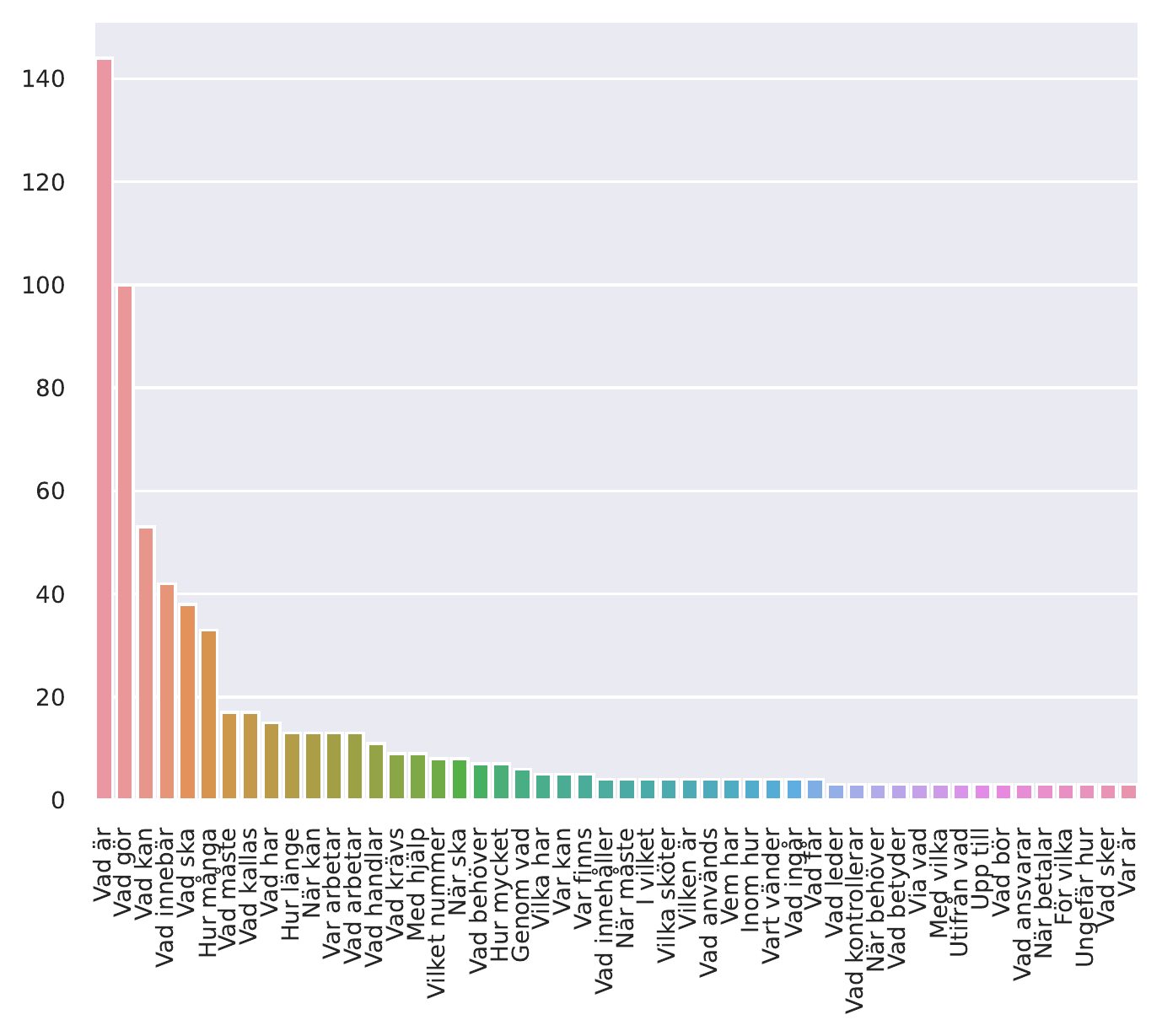}
	\vspace{-10px}
	\caption{The distribution of the first two words in questions of the training set of SweQUAD-MC}
	\label{fig:qword_stats_train}
\end{figure}

The dataset is relatively small with the training set consisting of 962 MCQs, the development (dev) set -- of 126 MCQs and the test set -- of 102 MCQs. The distribution of the first two words\footnote{which is a decent proxy for question words} in questions in the training set is shown in Figure \ref{fig:qword_stats_train}.

\section{Method}
We used Quinductor \cite{kalpakchi2021quinductor}, which is a mostly deterministic data-driven method applicable to any language having a dependency parser based on Universal Dependencies \citep{nivre2020universal}. In particular, it is applicable to Swedish. Quinductor is only capable of inducing QA-pairs based on single declarative sentences, which is suitable for assessing RC on the information-locating level. The method is data-driven, requiring a corpus of texts with the associated QA-pairs as training material, as well as some additional files, detailed in Appendix \ref{app:A}.

\begin{figure}[!hbt]
	\centering
	\includegraphics[width=0.49\textwidth]{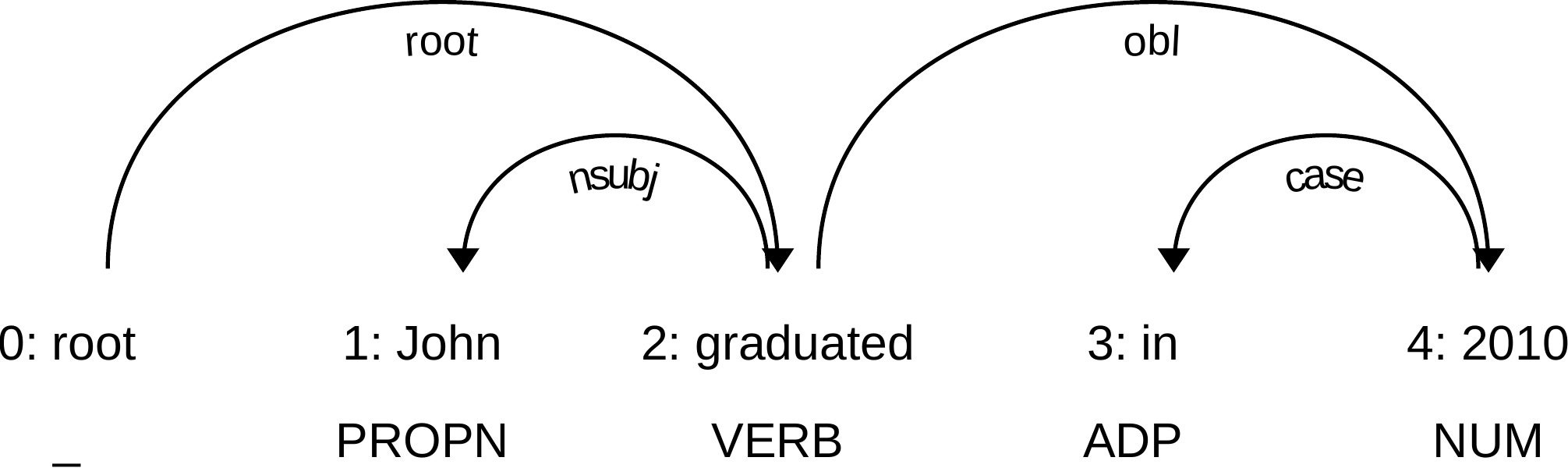}
	\caption{The dependency tree for the sentence ``John graduated in 2010''}
	\label{fig:quinductor_ex}
\end{figure}

\begin{figure}[!hbt]
	\centering
	\includegraphics[width=0.49\textwidth]{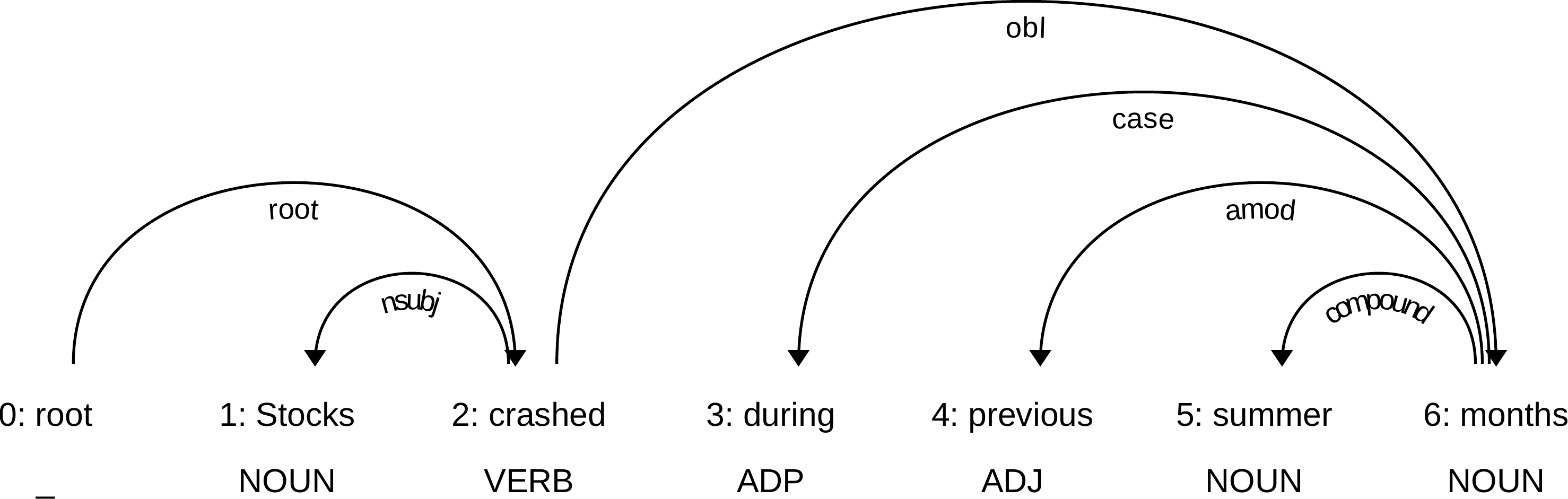}
	\caption{The dependency tree for the sentence ``Stocks crased during previous summer months''}
	\label{fig:quinductor_ex2}
\end{figure}

Quinductor works in two stages: first inducing the templates and then using them to generate questions from unseen sentences. In short, the first stage involves learning from data to express each given QA-pair in terms of the dependency structures of the source sentence, which is a basis for both the question and the answer. To exemplify, consider the sentence ``John graduated in 2010'' (with its dependency tree in Figure \ref{fig:quinductor_ex}) and the question ``When did John graduate?'' with the answer being in ``2010''. Quinductor will then learn to induce the template (\ref{ex1_q}) for the question and the template (\ref{ex1_a}) for the answer.

\enumsentence{\begin{tabbing}
		\texttt{When did} \=\texttt{[r.nsubj\#1]} \=\texttt{[r.lemma]} \texttt{?}\\
		\>~~~~~~~~~~{\small (John)}
		\>~~~~{\small (graduate)} 
	\end{tabbing}\label{ex1_q}
}
\enumsentence{\begin{tabbing}
		\=\texttt{<r.obl\#4>}\\
		\>~~~~~{\small (in 2010)}
	\end{tabbing}\label{ex1_a}}
At test time, Quinductor applies an overgenerate-and-rank strategy, and attempts to apply all of the induced templates to each given sentence. Clearly, the more sentences with similar dependency structures will be present in unseen data, the more successful the method will be. Also, the more template expressions with angled brackets (matching a whole phrase, as in (\ref{ex1_a}) above) are present in a template, the higher the generalization chance is. To exemplify, if we get a new sentence ``Stocks crashed during previous summer months'' (with its dependency tree in Figure \ref{fig:quinductor_ex2}), our previously induced template will be able to fire and produce the QA-pair ``When did stocks crash?'' -- ``during previous summer months'', although the trees are clearly not identical. For further details on the method and its generalization capabilities we refer to the original article.

\begin{table}[!t]
	\centering
	\begin{tabular}{p{4.5cm}c}
		\midrule
		\textbf{Statistics} & \textbf{Value}\\
		\midrule
		Number of induced templates & 248 \\
		Support per template &   \\
		\FirstIndent Mean $\pm$ STD & $1.04 \pm 0.23$ \\
		\FirstIndent Median (Min - Max) & 1 (1 - 3)\\ 
		\bottomrule
	\end{tabular}
	\caption{\label{tab:alg-prop} Descriptive statistics of the templates produced using the training set of the SweQUAD-MC dataset. Support per template is the number of sentences from the training set that yield the same template.}
\end{table}

In this work we have induced templates based on the training set of SweQUAD-MC (see more implementation details in Appendix \ref{app:A}). As can be seen in Table \ref{tab:alg-prop}, this resulted in 248 templates with most of them being unique, i.e. induced from (being supported by) only one sentence from the training set. Some templates had higher support with the maximum of 3 sentences per template.

\section{Evaluation and discussion}
\label{sec:eval}
The proportion of source sentences (the ones where the correct answer is found) from the dev. and test sets of SweQUAD-MC, for which Quinductor could generate something is reported in Table \ref{tab:eval-prop}.

\begin{table}[!b]
	\centering
	\begin{tabular}{p{4.5cm}cc}
		\midrule
		\textbf{} & \textbf{dev} & \textbf{test} \\
		\midrule
		\# of questions in the set & 126 & 102 \\
		\# of generated questions & 207 & 213 \\
		\midrule
		(1) SS with $\ge$ 1 applicable template & 64 & 44 \\
		(2) SS with $\ge$ 1 generated question after basic filtering & 49 & 36\\
		(3) SS with $\ge$ 1 generated question after mean filtering & 29 & 24\\
		(3) as \% of the respective set & 23\% & 23.5\% \\
		\midrule
		Generated questions per SS &  &  \\
		\FirstIndent Mean & 3.2 & 4.8 \\
		\FirstIndent Standard deviation & 4.1 & 7.2 \\
		\FirstIndent Median & 2 & 3 \\
		\FirstIndent Minimum & 0 & 0 \\
		\FirstIndent Maximum & 23 & 46 \\
		\bottomrule
	\end{tabular}
	\caption{\label{tab:eval-prop} Descriptive statistics of the questions induced on the dev. and test sets of the SweQUAD-MC using the templates, mentioned in Table \ref{tab:alg-prop}. ``SS'' stands for ``source sentence(s)'', i.e., the sentence in which the correct answer is found. ``$\ge 1$ applicable template'' means that at least 1 question was induced from the given SS.}
\end{table}

For evaluation, we took all 29 QA-pairs generated for the development set and all 24 QA-pairs generated for the test set of SweQUAD-MC. These 53 QA-pairs were combined with 53 original QA-pairs corresponding to the same source sentences from the respective corpora (later referred to as {\em gold\/} QA-pairs). The resulting 106 QA-pairs and the corresponding source sentences formed 106 evaluation triples, and were presented to 2 human judges (one native Swedish speaker and one non-native, but with a high proficiency) in a random order (different for each judge). Following \citet{kalpakchi2021quinductor}, we required judges to evaluate each triple using a questionnaire consisting of 9 criteria. Each criterion required a judgement on a 4-point Likert-type scale sfrom 1 (``Disagree'') to 4 (``Agree''). The evaluation itself was conducted online on an in-house instance of Textinator Surveys \cite{kalpakchi-boye-2022-textinator}. The evaluation guidelines are reported in Appendix \ref{app:eval}.

Five criteria concerned questions and were formulated as statements asking whether a question: 
\begin{enumerate}
	\item[C1] is grammatically correct $(\uparrow)$
	\item[C2] makes sense $(\uparrow)$
	\item[C3] would be clearer if more information were provided $(\downarrow)$
	\item[C4] would be clearer if less information were provided $(\downarrow)$
	\item[C5] is relevant to the given sentence $(\uparrow)$
\end{enumerate}
The remaining 4 criteria concerned the answer and asked whether the suggested answer: 
\begin{enumerate}
	\item[C6] correctly answers the question $(\uparrow)$
	\item[C7] would be clearer if phrased differently $(\downarrow)$
	\item[C8] would be clearer if more information were provided $(\downarrow)$
	\item[C9] would be clearer if less information were provided $(\downarrow)$
\end{enumerate}
$\uparrow$ ($\downarrow$) indicates that the higher (lower) the judgements on the Likert scale, the better.

\begin{table}[!t]
\centering
\begin{tabular}{lccccc}
\midrule
\multirow{2}{*}{\textbf{Criterion}} & \multirow{2}{*}{} & \multicolumn{2}{c}{\textbf{dev}} & \multicolumn{2}{c}{\textbf{test}} \\
& & \textbf{gold} & \textbf{gen} & \textbf{gold} & \textbf{gen}\\
\midrule
\multirow{2}{*}{\parbox{1cm}{C1 $\uparrow$}} & $\kappa$ & 0.86 & 0.54 & 0.83 & 0.33 \\
& $\gamma$ & 0.92 & 0.78 & 0.91 & 0.55 \\
\midrule
\multirow{2}{*}{\parbox{1cm}{C2 $\uparrow$}} & $\kappa$ & 0.59 & 0.49 & 0.88 & 0.39 \\
& $\gamma$ & 0.83 & 0.79 & NA/4 & 0.75 \\
\bottomrule
\end{tabular}
\caption{Inter-annotator agreement for criteria C1 and C2 on the development and test sets of SweQUAD-MC. $\gamma=\text{NA/X}$ means that at least one annotator gave the same score X to all evaluation triples, so it is impossible to count concordant and discordant pairs.}
\label{tab:human-eval-agreement}
\end{table}

Following \citet{kalpakchi2021quinductor}, we have measured IAA using Randolph's $\kappa$ \cite{randolph2005free} and Goodman-Kruskall's $\gamma$ \citep{goodman1979measures}. The former, $\kappa$, ranging between $-1$ and $1$, accounts for agreement in absolute rankings, i.e. being boosted if the annotators gave exactly same score to an evaluation triple. The value of 0 indicates the level of agreement that could be expected by chance, the positive (negative) values indicate agreement better (worse) than chance. The latter, $\gamma$, also ranging between $-1$ and $1$, accounts for agreement in relative ordering, i.e. being boosted if the annotators ordered a pair of two evaluation triples in the same way, no matter the actual scores. $\gamma=0$ indicates no agreement, $\gamma=1$ denotes a complete agreement and $\gamma=-1$ hints at a perfect disagreement. 

%Among gold
%On the dev set, the annotators found all 29 to be grammatically correcct, 26 questions out of 29 to make sense and the same 26 questions -- both to be grammatically correct and to make sense.
%On the test set, the annotators found 23 questions out of 24 to be grammatically correct (median is 3 or higher), 23 questions out of 24 to make sense and the same 23 questions -- both to be grammatically correct and make sense.
%
%Among the generated
%On the dev set, the annotators found 17 questions out of 29 to be grammatically correcct, 11 questions out of 29 to make sense and the same 11 questions -- both to be grammatically correct and to make sense.
%On the test set, the annotators found 10 questions out of 24 to be grammatically correct (median is 3 or higher), 8 questions out of 24 to make sense and 7 questions -- both to be grammatically correct and make sense.

Without a doubt, the two basic criteria, which must necessarily have high judgements in order to even consider looking at all other criteria, are C1 and C2. Table \ref{tab:human-eval-agreement} shows that the evaluators had high agreement on both criteria, especially when it comes to relative ordering of triples ($\gamma > 0.5$).

\begin{figure}[!t]
	\centering
	\begin{subfigure}{0.36\textwidth}
		\includegraphics[width=\textwidth]{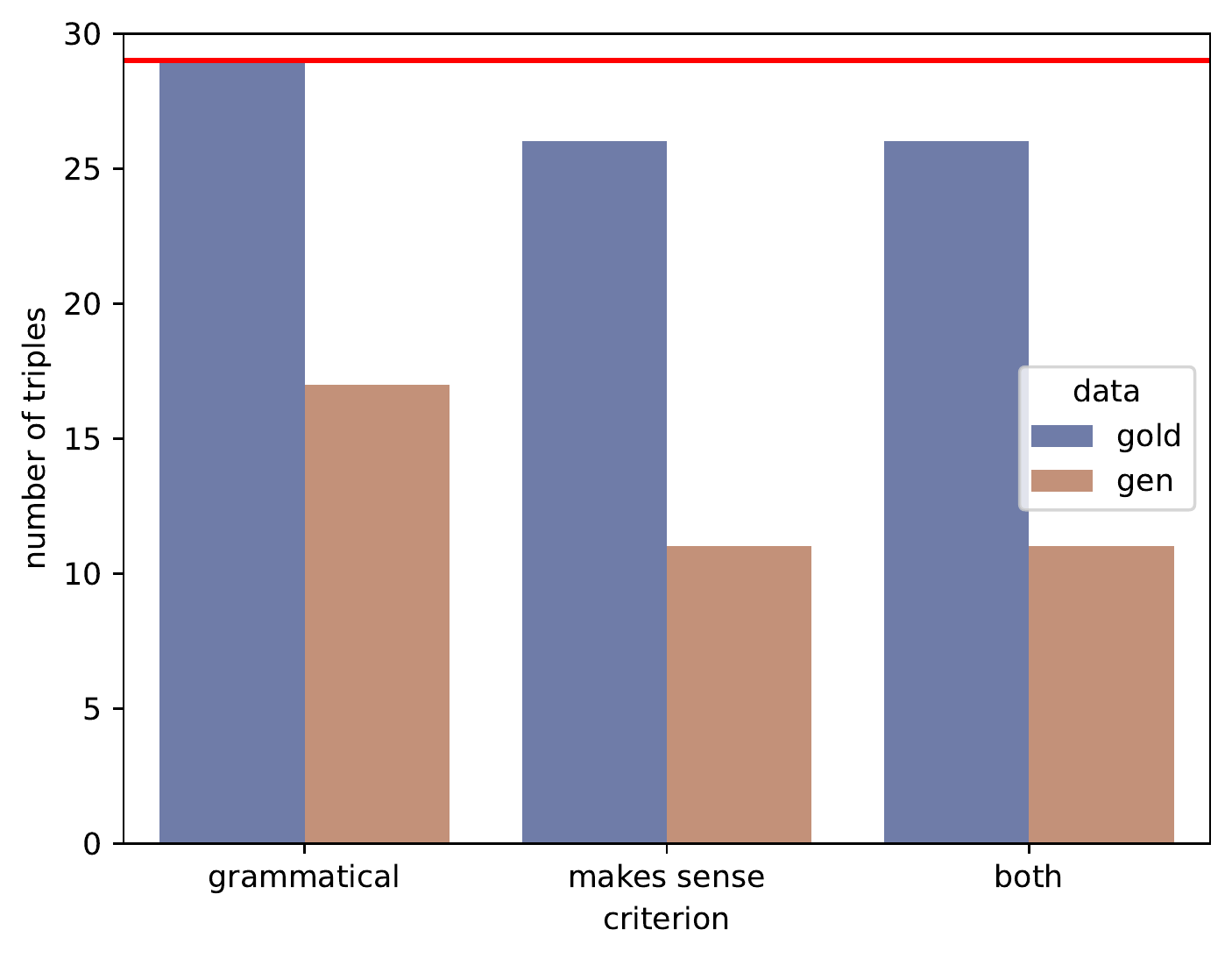}
		\vspace{-22px}
		\subcaption{Development set}
		\label{fig:dev_gns}
	\end{subfigure}
	
	\smallskip
	\begin{subfigure}{0.36\textwidth}
		\includegraphics[width=\textwidth]{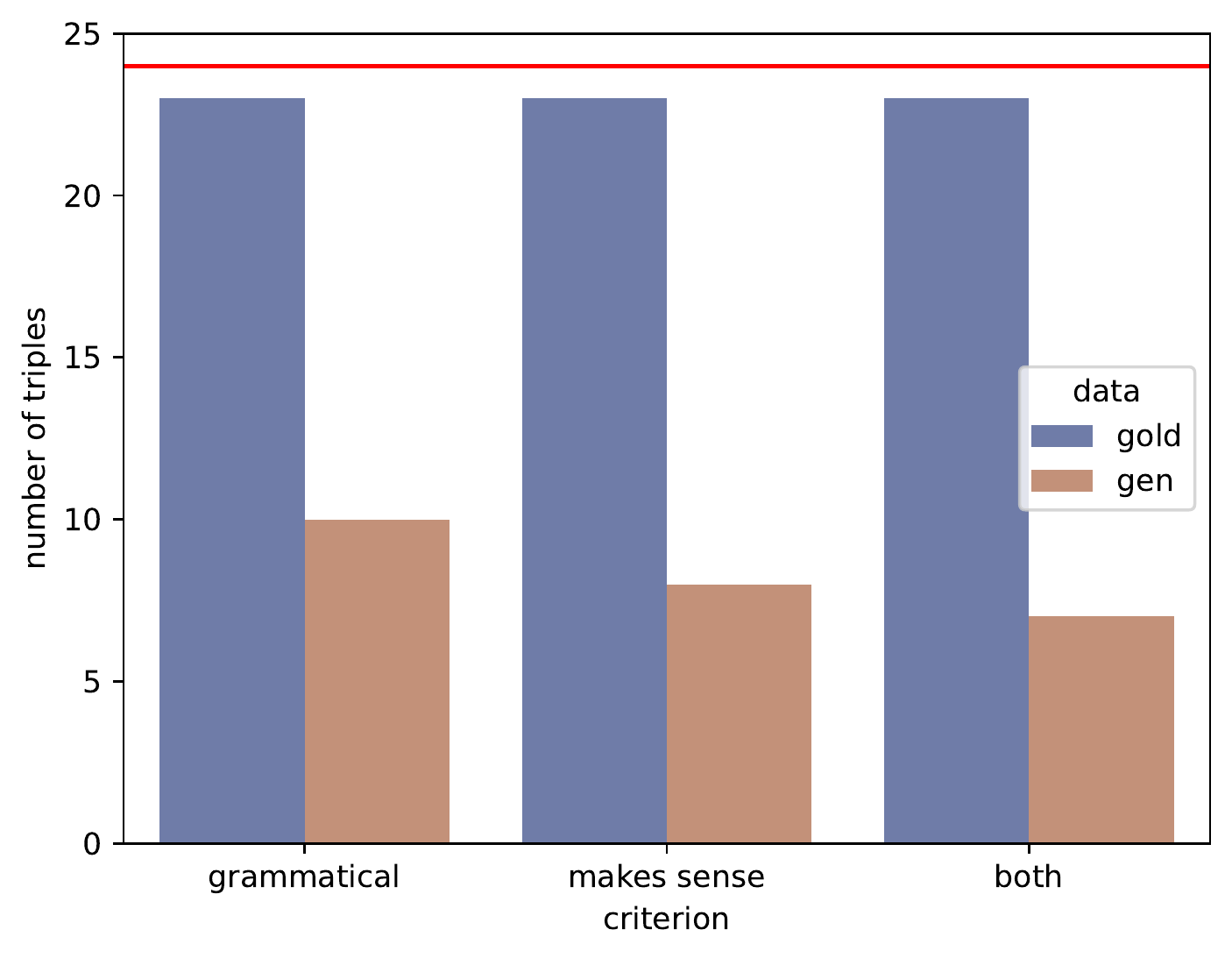}
		\vspace{-22px}
		\subcaption{Test set}
		\label{fig:test_gns}
	\end{subfigure}
	
	\vspace{-5px}
	\caption{Barplots of the number of evaluation triples with rated highly (median $\geq$ 3) on C1 (grammatical), C2 (makes sense), and both C1 and C2. The horizontal red lines indicate a total number of evaluation triples that were considered (equal for gold and generated data)}
	\label{fig:gns}
\end{figure}

As can be seen in Figure \ref{fig:gns}, nearly all \emph{gold} questions of the dev and test sets were judged highly on C1 (median $\geq 3$). At the same time, 17 ($\sim58\%$) and 10 ($\sim42\%$) of the \emph{generated} questions of the dev and test sets respectively were rated highly on C1, resulting in a total of $50.9\%$ between the sets.

Figure \ref{fig:gns} also reveals that surprisingly 4 \emph{gold} questions in total between the dev and test sets were judged as \textbf{not} making sense (median $< 3$). At the same time only 11 ($\sim38\%$) and 8 ($\sim33\%$) of the \emph{generated} questions of the dev and test sets respectively were rated highly on C2 (median $\geq 3$), resulting in a total of $35.8\%$ between the sets. We also observed that only one generated question was judged highly on C2, but lower on C1.

% Grammatical, makes sense and relevant for sentence
%    dev: 11, test: 6
% Grammatical, makes sense and suitable CA
%    dev: 7, test: 5
% Grammatical, makes sense, relevant for sentence and suitable CA (ultimate)
%    dev: 7, test: 5

We analyzed further only 19 \emph{generated} QA-pairs between the sets that were judged highly on both C1 and C2. The analysis was carried out in terms of criteria C5 and C6, the agreement on both of which was reasonably high (see Appendix \ref{app:E}). Out of these 19, 17 (11 on the dev set and 6 on the test set), were also rated highly (median $\geq 3$) on C5 and 12 ($63.2\%$), 7 on the dev set and 5 on the test set, were rated highly on C6. Crucially, there were \textbf{no} QA-pairs with the suggested answer being judged as correct (high median rating on C6), but with the question being deemed as irrelevant to the given sentence (low median judgement on C5). This means that out of all generated QA-pairs, 12 ($22\%$) were judged as completely valid, 7 ($24\%$) on the dev set and 5 ($21\%$) on the test set, making Quinductor a resonably strong RC-QG baseline for SweQUAD-MC. 

Some examples of successful generation are presented in Appendix \ref{app:B}. Further analysis (in particular of errors) and related generation examples are presented in Appendix \ref{app:C}. Although automatic evaluation metrics provide very limited insights, we also report them in Appendix \ref{app:D}, following \cite{kalpakchi2021quinductor}.

%Questions to be answered:
%1. How many questions of the gold and gen were found to be grammatical and making sense. -- DONE
%2. Out of these and for generated only, how many were (a) relevant, (b) had correct correct answers
%3. Which problems did the generated QA-pairs have and to what extent?

\section*{Acknowledgments}

This work was supported by Vinnova (grant 2019-02997), and Digital Futures (project SWE-QUEST).

\bibliographystyle{acl_natbib}
\bibliography{emnlp2020}

\appendix

\section{Implementation details}
\label{app:A}
We have used dependency parsing models from the package called Stanza\footnote{\url{https://github.com/stanfordnlp/stanza}} \cite{qi-etal-2020-stanza}, version 1.4.2 and used quinductor\footnote{\url{https://github.com/dkalpakchi/quinductor}} package \cite{kalpakchi2021quinductor}, version 0.2.2. This is important, because it is likely that the induced templates will differ if different models are used. The supplementary files necessary for either generating templates or during ranking the generated questions were obtained as follows:
\begin{itemize}
	\item the IDFs were calculated based on the SweQUAD-MC training set (necessary for inducing templates);
	\item the morphological n-gram model was calculated based on training, dev and test sets of the UD's Talbanken treebank\footnote{\url{https://universaldependencies.org/treebanks/sv_talbanken/index.html}} (necessary for ranking);
	\item the question-word model was calculated based on the SweQUAD-MC training set (necessary for ranking).
\end{itemize}
Note that the question-word model was induced using one of the earlier versions of Stanza, which is why the distributions differ slightly, compared to the 1.4.2. All these suplementary files are released in the GitHub repository, associated with this paper.

\section{Human evaluation guidelines}
\label{app:eval}
Here we report the exact wording of the guidelines presented to the human evaluators.
\begin{tcolorbox}
Tack att du deltar i vår utvärdering av läsförståelsefrågor! Du kommer att se ett antal meningar (en i taget) tillsammans med fråga och det rätta svaret (QA-par). Du kommer också att se ett antal påstående för varje QA-par. Din uppgift är att bestämma i vilken utsträckning du håller med varje påstående.

Om frågan är obegriplig ska du välja "1" för alla påståenden relaterade till det föreslagna svaret.

Vänligen ignorera alla möjliga formatteringsfel, t.ex. skiljetecken (.,!?:;) eller versaler som saknas.
\end{tcolorbox}

\section{Generation examples}
\label{app:B}

\qgnumsample{1}{inom h\"also- och sjukv\r{a}rden arbetar dietisten med nutritionsbehandling och kostr\r{a}dgivning, b\r{a}de med enskilda patienter och i grupp.}{var arbetar dietisten med nutritionsbehandling och kostr\r{a}dgivning?}{inom h\"also- och sjukv\r{a}rden}

\qgnumsample{2}{om du \"ar borta l\"angre \"an ett \r{a}r eller om du planerar att bos\"atta dig i ett annat land kan migrationsverket \r{a}terkalla ditt uppeh\r{a}llstillst\r{a}nd.}{vad kan migrationsverket g\"ora om du \"ar borta l\"angre \"an ett \r{a}r eller om du planerar att bos\"atta dig i ett annat land?}{\r{a}terkalla uppeh\r{a}llstillst\r{a}nd}

\qgnumsample{3}{hovslagaren behandlar \"aven skador i hoven till exempel hovsprickor eller hovb\"older.}{vilka behandlar hovslagaren?}{skador i hoven}

\qgnumsample{4}{pulverf\"arg appliceras ofta p\r{a} metallkomponenter.}{vad appliceras ofta på metallkomponenter?}{pulverf\"arg}

\qgnumsample{5}{milj\"o- och h\"alsoskyddsinspektörer samarbetar till exempel med andra tj\"anstemän inom kommuner, f\"oretag och myndigheter.}{vad g\"or milj\"o- och h\"alsoskyddsinspekt\"orer?}{samarbetar till med andra tj\"anstem\"an inom kommuner , f\"oretag och myndigheter}

\section{Error analysis}
\label{app:C}
The QA-pairs that are rated low on C2 (question makes sense) represent the worst kind of errors, which are impossible to correct automatically (like, say, some grammatical errors). Hence, let's examine closer some examples of evaluation triples with generated QA-pairs containing questions that were rated low (median $< 3$) on C2. Please ignore formatting errors (like absense of punctuation or capital letters).

The questions from the following triples got a median score of 1 on C2:
\qgnumsample{1}{arbetsuppgifterna varierar beroende på var man är anställd men den främsta uppgiften är att se till så att hyresgästerna känner sig trygga och nöjda i sin arbets- och boendemiljö.}{vad gör arbetsuppgifterna beroende på var man är anställd men den främsta uppgiften är att se till så att hyresgästerna känner sig trygga och nöjda i sin arbets- och boendemiljö?}{varierar}
\qgnumsample{2}{du kan då behöva vänta på ditt nya beslut utanför sverige.}{vad kan du göra på ditt nya beslut utanför sverige?}{vänta}

The questions from the following triples got a median score of 1.5 on C2:
\qgnumsample{3}{den största delen består av barrskog, men i söder breder lummig lövskog ut sig.}{vad består den största delen från?}{barrskog}
\qgnumsample{4}{en stuntman ersätter filmskådespelare i riskfyllda situationer.}{var ersätter stuntman?}{i riskfyllda situationer} 

The question from the following triple got a median score of 2 on C2:
\qgnumsample{5}{till exempel tillbringar kirurgen en stor del av sin tid i operationssalen, medan en psykiatriker arbetar mycket med samtal med sina patienter.}{vem tillbringar till en stor del av sin tid i operationssalen medan en psykiatriker arbetar mycket med samtal med sina patienter?}{kirurgen}

The questions from the following triples got a median score of 2.5 on C2:
\qgnumsample{6}{en sjukskriven medarbetare kostar lågt räknat 2 740 kronor per dag*.}{vad gör en sjukskriven medarbetare?}{kostar kronor}
\qgnumsample{7}{man bör ha god fysik då arbetet kan slita på nacke och axlar.}{vad bör man göra då arbetet kan slita på nacke och axlar?}{ha fysik}

Question 1 was generated using the template \texttt{vad g\"or [r.nsubj\#1] <r.advcl\#2>?}. The problematic one in the case of sentence 1 turned out to be the last template expression \texttt{<r.advcl\#2>}, which takes the whole subtree, whose root can be found by following the arc labeled \texttt{advcl} from the root \texttt{r} of the dependency tree. In this case the subtree was too large and thus the generated question contains a lot of unnecessary information. This is one typical type of errors that could be referred to as \emph{overgeneralization}, i.e. the case when the induced templates become too general. However, a balance between letting Quinductor to induce too generic or too specific templates is difficult to strike, so such kinds of errors are inevitable. Question 5 suffers from the same problem.

Question 2 is based on the template \texttt{vad [r.aux\#1] [r.nsubj\#2] g\"ora <r.obl\#3>?}, which in turn was generated from the sentence ``vad beh\"over ambulanssjuksk\"oterskan g\"ora vid st\"orre olyckor med m\r{a}nga skadade?''. The structure is clearly very similar, but the preposition of the oblique nominal happened to be different, so the template expression \texttt{<r.obl\#3>} did not generalize correctly. In fact questions 3 suffers from similar problems.

Question 4 represents the case when the question word from the template is wrong. All question words are guaranteed to be recorded verbatim as strings and not deduced from the dependency tree of the sentence, so such errors are also inevitable in the current version of Quinductor.

QA-pairs 6 and 7 represent more successful applications of templates for the questions, although the questions are still not perfectly intelligble. The major problem with these QA-pairs are the answers, which in fact suffer an inverse problem compared to that of question 1. For instance, the template for the suggested answer for question 7 is \texttt{[r] [r.obj\#4]}. The problem with this template is that it just picks up only specific nodes of the dependency tree and is prone to errors if the object, referred by \texttt{r.obj\#4} will turn out to have some dependents, which will also be relevant to include in the answer. This problem could be referred to as \emph{undergeneralization}. As previously mentioned, the balance between over- and undergeneralization is hard to strike, especially given that Quinductor is a data-driven method and all templates do depend on the data at hand.

We noted that sometimes these errors arise, because of inconsistency of dependency parsers themselves, which is not a new observation. For instance, \citet{kalpakchi-boye-2021-minor} showed that even changes as minor as replacing one 4-digit numeral by another can cause surprisingly large inconsistencies in the resulting trees using the state-of-the-art dependency parsers, in particular for Swedish.

\section{Automatic evaluation metrics}
\label{app:D}
Following \cite{kalpakchi2021quinductor}, we have calculated BLEU-N, ROUGE-L and CIDEr using nlg-eval \cite{sharma2017nlgeval} and METEOR using METEOR-1.5 \cite{denkowski2014meteor} and specifying the language to be Swedish.
\begin{table}[!htb]
	\centering
	\begin{tabular}{lcc}
		\midrule
		\textbf{Metric} & \textbf{dev} & \textbf{test} \\
		\midrule
		BLEU-1 & 0.39 & 0.22 \\
		BLEU-2 & 0.29 & 0.15 \\
		BLEU-3 & 0.22 & 0.09 \\
		BLEU-4 & 0.18 & 0.06 \\
		METEOR & 0.33 & 0.21 \\
		ROUGE-L & 0.38 & 0.27 \\
		CIDEr & 0.76 & 0.84 \\
		\bottomrule
	\end{tabular}
	\caption{\label{tab:auto-eval} Automatic evaluation on the development and test sets of SweQUAD-MC only for generated questions ranked first.}
\end{table}

\section{Detailed IAA analysis}
\label{app:E}
Inter-annotator agreement (IAA) for criteria C3 - C9 is presented in Table \ref{tab:human-eval-agreement-2}. Let's start by analyzing the agreement on the development set. We observed that agreement on C5 - C8 is quite strong for both gold and generated QA-pairs. On the other hand, $\gamma$ is strongly negative for C3 (the question would be learer if more information were provided), indicating that the annotators relative ranking are almost exactly opposite. At the same time $\kappa$ on C3 is quite weak as well, especially on the generated QA-pairs. For the generated questions both $\kappa = 0.08$, so $\gamma = -0.7$ indicate almost no agreement in absolute scores and a strong disagrement in relative orderings.

\begin{table}[!t]
	\centering
	\begin{tabular}{lccccc}
		\midrule
		\multirow{2}{*}{\textbf{Criterion}} & \multirow{2}{*}{} & \multicolumn{2}{c}{\textbf{dev}} & \multicolumn{2}{c}{\textbf{test}} \\
		& & \textbf{gold} & \textbf{gen} & \textbf{gold} & \textbf{gen}\\
		\midrule
		\multirow{2}{*}{\parbox{1cm}{C3 $\downarrow$}} & $\kappa$ & 0.45 & 0.08 & 0.83 & 0.17 \\
		& $\gamma$ & -1.0 & -0.7 & NA/1 & -0.09\\
		\midrule
		\multirow{2}{*}{\parbox{1cm}{C4 $\downarrow$}} & $\kappa$ & 0.95 & 0.91 & 1.0 & 0.83 \\
		& $\gamma$ & NA/1 & -1.0 & NA/1 & 1.0 \\
		\midrule
		\multirow{2}{*}{\parbox{1cm}{C5 $\uparrow$}} & $\kappa$ & 0.77 & 0.72 & 0.67 & 0.44 \\
		& $\gamma$ & 0.82 & 0.93 & NA/4 & 0.81 \\
		\midrule
		\multirow{2}{*}{\parbox{1cm}{C6 $\uparrow$}} & $\kappa$ & 0.72 & 0.63 & 0.67 & 0.56 \\
		& $\gamma$ & 0.79 & 0.88 & -1.0 & 0.78\\
		\midrule
		\multirow{2}{*}{\parbox{1cm}{C7 $\downarrow$}} & $\kappa$ & 0.63 & 0.49 & 0.94 & 0.83\\
		& $\gamma$ & 0.62 & 1.0 & NA/1 & 1.0\\
		\midrule
		\multirow{2}{*}{\parbox{1cm}{C8 $\downarrow$}} & $\kappa$ & 0.54 & 0.63 & 0.94 & 0.83\\
		& $\gamma$ & 0.4 & 0.73 & NA/1 & 0.93\\
		\midrule
		\multirow{2}{*}{\parbox{1cm}{C9 $\downarrow$}} & $\kappa$ & 1.0 & 1.0 & 1.0 & 1.0\\
		& $\gamma$ & NA/1 & NA/1 & NA/1 & NA/1\\
		\bottomrule
	\end{tabular}
	\caption{Inter-annotator agreement for criteria C3 - C9 on the development and test sets of SweQUAD-MC. $\gamma=\text{NA/X}$ means that at least one annotator gave the same score X to all evaluation triples, so it is impossible to count concordant and discordant pairs.}
	\label{tab:human-eval-agreement-2}
\end{table}

The scores for the gold questions on C3 are much more peculiar, namely $\gamma = -1.0$, while $\kappa = 0.45$, indicating complete opposite relative ordering between the annotators, while also having a moderate agreement in absolute ranks. If we look closer at the data, it turns out that annotator A gave almost all gold triples the score of 1 on C3, except one triple (let's call it $T_x$) that got the score of 3. Hence, this is the only triple that can be used to establish relative ordering for the annotator A. Now that very same $T_x$ got the score of 1 on C3 from the annotator B. This means that $T_x$ will always be scored higher than all other triples for the annotator A, whereas it will be scored lower than other triples for the annotator B. This, in turn, means that all relative orderings are opposite between the annotators, resulting in $\gamma = -1$. This example is a good illustrataion of why only $\kappa$ or $\gamma$ is not enough for assessing the IAA, but indeed both are needed and should be interpreted with caution. A similar situation is observed for the generated questions from the development set on the criterion C4, and for the gold questions from the test set on the criterion C6.

\end{document}